\documentclass[10pt,twocolumn,letterpaper]{article}

\usepackage{wacv}
\usepackage{times}
\usepackage{epsfig}
\usepackage{graphicx}
\usepackage{amsmath}
\usepackage{amssymb}

\usepackage{nicefrac} 
\usepackage{amsfonts} 
\usepackage{booktabs}
\usepackage{pifont}
\usepackage{breqn}

\usepackage{hyperref}
\usepackage{cite}

\DeclareMathOperator*{\argmin}{arg\,min}

\newcommand{\cmark}{\ding{51}}%
\newcommand{\xmark}{\ding{55}}%

\wacvfinalcopy 

\def\wacvPaperID{1082} 

\ifwacvfinal\fi
\begin{document}

\title{Unbiased Evaluation of Deep Metric Learning Algorithms}

\author{Istv\'an Feh\'erv\'ari \\
Amazon\\
{\tt\small istvanfe@amazon.com}
\and
Avinash Ravichandran \\
AWS\\
{\tt\small ravinash@amazon.com}
\and
Srikar Appalaraju \\
Amazon\\
{\tt\small srikara@amazon.com}
}

\maketitle
\ifwacvfinal\thispagestyle{empty}\fi

\begin{abstract}
   Deep metric learning (DML) is a popular approach for images retrieval, solving verification (same or not) problems and addressing open set classification. Arguably, the most common DML approach is with triplet loss, despite significant advances in the area of DML. Triplet loss suffers from several issues such as collapse of the embeddings, high sensitivity to sampling schemes and more importantly a lack of performance when compared to more modern methods. We attribute this adoption to a lack of fair comparisons between various methods and the difficulty in adopting them for novel problem statements.
   
   In this paper, we perform an unbiased comparison of the most popular DML baseline methods under same conditions and more importantly, not obfuscating any hyper parameter tuning or adjustment needed to favor a particular method. We find, that under equal conditions several older methods perform significantly better than previously believed. In fact, our unified implementation of 12 recently introduced DML algorithms achieve state-of-the art performance on CUB200, CAR196, and Stanford Online products datasets which establishes a new set of baselines for future DML research. The codebase and all tuned hyperparameters will be open-sourced for reproducibility and to serve as a source of benchmark.
\end{abstract}

\section{Introduction}
The goal of metric learning is to learn a function that maps an image to a high-dimensional vector embedding space such that the representation of semantically similar images are closer together while the representation of dissimilar images are farther away. Such functions allow efficient clustering \cite{37732}, visual search \cite{Jing:2015:VSP:2783258.2788621}, recommendations \cite{lee2018collaborative}, and few-shot learning \cite{snell2017prototypical} amongst other applications \cite{istvanwacv19}. In this paper, we explore deep neural network model as a function approximator for metric learning. 

\begin{figure}[t]
	\centering
	\includegraphics[width=\linewidth]{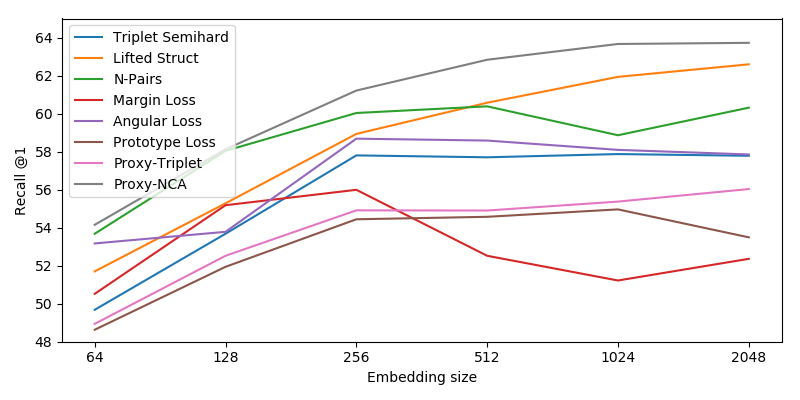}
	\caption{Retrieval performance with different embedding sizes on CUB200 with Inception-BN (best viewed in color).}
	\label{fig:embeddingsize}
\end{figure}

\begin{figure}[t]
	\centering
	\includegraphics[width=\linewidth]{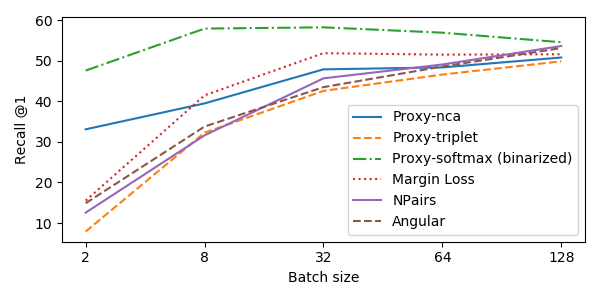}
	\caption{Retrieval performance with different batch sizes on CUB200 with Inception-BN (best viewed in color).}
	\label{fig:batchsize}
\end{figure}

Let $f:\Omega \subseteq \mathbb{R}^{D_{i}}  \rightarrow \mathbb{R}^{D_{o}}$ be the differentiable DML function that maps an image $z \in \Omega$, the space of images to $f(z) \in \mathbb{R}^{D_{o}} $, a $D_{o}$-dimensional space. The goal is to enable $f(z_i)$ to learn to keep similar data points close and dis-similar points far in this $D_{o}$ dimensional space. The main challenge with metric learning is to sample the right set of images that provide sufficient information that will help the training move towards a point in the parameter space that has a lower loss value. In fact, when samples are very "easy", the gradients will be close to zero, preventing any parameter updates from taking place. On the other hand, very "hard" samples might destabilize the training process collapsing it to a singular embedding. 

In recent years, there have been many approaches to solve this issue -  data sampling strategies, different deep neural network architectures and various loss or distance functions. Data sampling strategies work by either  utilizing relationships within a batch \cite{sohn2016improved} or by implementing more effective sampling strategies such as \cite{wu2017sampling} to sample mini-batches. The drawback of these approaches is that they are usually compute intensive and need a large batch size to work efficiently which is typically constrained by the underlying hardware. Furthermore, as these sampling strategies employ several heuristics, it is hard to pick a sampling strategy for novel problems.

Deep neural network architectures for image feature extraction have seen great improvements over the last years as well. Consequently, most DML methods are only measured with the state-of-the-art network architecture which makes comparisons to older methods rather challenging given that newer architectures are expected to deliver better results out of the box.

Finally, while we seek better loss functions to improve the field, often these new losses do not work as they are claimed across datasets. This is partly due to the fact that different losses require different embedding properties. However, it is not yet fully understood how the performance of a loss function depends on the size of the embedding.

Another recent promising direction is to move towards parametric learning via proxy embeddings where neither sampling, nor high batch sizes are necessary \cite{movshovitz2017no}\cite{normproxies} at the cost of extra (constant) memory during training. These methods are often criticized for their lack of scalability given the best performing proxy-based methods require at least one dedicated embedding for each class in the training set.

Unfortunately, most papers evaluate their approaches under different conditions such as bigger embedding size, better feature extractor, or by using additional information (e.g. bounding boxes, text modalities) while claiming no influence of these on the results. Table \ref{experiment-variety} presents an overview of the diversity of settings that are used by various algorithms. 
In this paper, we hypothesize that available comparisons of popular DML methods are done improperly concealing the true order of performance of these methods. Therefore, we re-evaluate the most prominent baselines under same conditions to provide a more reliable comparison on three popular datasets: the CUB-200-2011 \cite{WahCUB_200_2011}, the CARS-196 \cite{cars196}, and the Stanford Online Products \cite{oh2016deep}. In particular, we study the effects of different embedding sizes and the two most popular CNN feature extractors with respect to various loss functions and sampling strategies. Surprisingly, we find that several methods perform a lot better than expected under the right set of configurations, while some novel methods significantly underperfom when the comparison is more fair. Furthermore, while attempting to reproduce the original results of respective papers, we uncovered a set of previously unmentioned tricks that are imperative for obtaining state-of-the-art results. Our findings conclude that triplet loss even with semi-hard mining performs the worst in retrieval and clustering tasks among all the methods we tested.

The main contributions of this paper are the following:
\begin{itemize}
	\item We provide a concise survey of modern deep metric learning approaches and compare them under the same constraints.
	\item We show that the performance ranking of the tested methods do not follow the expected order once the conditions are equal.
	\item Our findings show that methods that optimize directly for clustering quality perform overall surprisingly better than previously expected.
	\item We analyze the effect of loss functions, different embedding and batch sizes, and two different feature extractors across all methods.
	\item We release\footnote{https://github.com/ifeherva/DMLPlayground} our implementations of 12 different state-of-the-art algorithms in MXNet\footnote{https://mxnet.apache.org/} which we use for comparison. This framework can be extended to new datasets and contains novel feature-extractor architecture combinations that have not appeared in literature. We discuss how our framework enabled easy experimentation with novel combinations to present new state-of-the-art on CUB200, CAR196 and Stanford products datasets.
\end{itemize}

\begin{table*}[]
	\centering
	\begin{tabular}{@{}llccc@{}}
				\toprule
		Method & Feature extractor & Embedding size & R@1 & Normalize last layer \\ \midrule
		Triplet Semihard  \cite{schroff2015facenet} & GoogLeNet & 64 & 42.6 & \xmark \\
		Lifted Struct  \cite{oh2016deep} & GoogLeNet & 64  & 43.6 & \xmark \\
		Npairs \cite{sohn2016improved} & GoogLeNet & 64   & 45.4 & \xmark \\
		Struct Clust \cite{oh2017deep} & Inception-BN  & 64  & 48.2 & \cmark \\
		Margin Loss \cite{wu2017sampling} & ResNet50 & 128 & 63.6 & \cmark   \\
		Angular Loss \cite{wang2017deep} & GoogLeNet & 512 & 53.6 & \xmark \\
		Prototype Loss \cite{snell2017prototypical} & GoogLeNet & $\text{1024}^{\dagger}$  & - & \xmark \\
		Proxy-Triplet \cite{movshovitz2017no} & Inception-BN & 64  & -  & \cmark\\
		Proxy-NCA \cite{movshovitz2017no} & Inception-BN & 64 & 49.2 & \cmark \\
		Proxy-Softmax \cite{normproxies} & Inception-BN & 512 & 55.3 & \cmark \\
		Ranked List Loss \cite{Wang_2019_CVPR} & Inception-BN & 512 & 57.4 & \cmark \\
		DREML \cite{Xuan_2018_ECCV} & GoogleNet & 12x48 & 58.9 & \xmark \\
	\end{tabular}
	\caption{Reported experimentation settings and recall @1 on the CUB200 dataset. $\dagger$: embedding is multimodal.}
	\label{experiment-variety}
\end{table*}

\section{Distance Metric learning}

The goal of deep metric learning is to learn a distance $d(\mathbf{x}_{a},\mathbf{x}_{b}, \theta)$ between two data points $\mathbf{x}_{a}, \mathbf{x}_{b}  \in  \mathbb{R^N}$ usually formulated as the Euclidean or cosine distance between embeddings processed via deep neural networks with parameters $\theta$. Training takes place in a supervised fashion where a set of similar and dissimilar points provide the similarity relationships in the image domain. For example, contrastive loss \cite{contrastiveloss} uses a pair of similar or dissimilar points with a single binary label encoding the similarity relationship. The main downside of this approach is that the optimization of the positive pairs is independent from the negative pairs, although the optimization should force the distance between positive pairs to be smaller than negative pairs. 

Triplet loss \cite{tripletloss} was introduced to address this issue, which is defined over three points: $T=\{(\mathbf{x}_{a},\mathbf{x}_{p},\mathbf{x}_{n})\}$, where $\mathbf{x}_{a}$ and $\mathbf{x}_{p}$ have the same label (called anchor and positive) and $\mathbf{x}_{n}$ have a different label (called negative). 
\begin{equation} \label{eq:triplet_loss}
L_{\text{triplet}}(\mathbf{x}_{a},\mathbf{x}_{p},\mathbf{x}_{n}) = \left [d(\mathbf{x}_{a},\mathbf{x}_{p}) - d(\mathbf{x}_{a},\mathbf{x}_{n}) + M \right ]_{+}
\end{equation}
where $\left [ \cdot \right ]$ is the hinge function and $M$ is the margin. Though triplet loss yields considerably higher performance than previous approaches, it suffers from the same issues as contrastive loss: its margin constant requires careful tuning. Furthermore, its runtime complexity is $\mathcal{O}(N^{3})$ which necessitates a mining strategy to find informative triplets during the training process. In practice, triplets that are considered easy waste computation and slow down convergence. On the other hand, sampling only hard triplets can easily destabilize the training process. There have been several proposals on how to solve this problem: either by changing the batch selection process, or by introducing novel loss functions that do not suffer from the drawback of hinge-based losses.
In the following we will review a (certainly not exhaustive and not chronological) list of the most popular approaches of the field; what aspect of the problem they are solving and what other challenges they are introducing. 

\subsection{Semi-hard sample mining}
A simple way to improve the convergence of triplet loss is to increase the batch size, thus improving the probability of sampling a useful triplet. However, large batch sizes are typically constrained by available GPU memory and will introduce extra computations. Several better sample selection strategies were proposed to mine for triplets that are most useful to the training. For example, \cite{RippelPDB15} proposes a sampling strategy based on neighbor classes. A more efficient solution was presented in \cite{schroff2015facenet}, where the semi-hard triplets were selected during training by examining the pairwise similarity between samples of the same batch. Arguably this is the most widely used algorithm in DML and thus can be used as a good baseline to compare against more advanced models.

\subsection{Distance weighted sampling and margin loss}
The importance of sampling has been shown in \cite{wu2017sampling} by sampling triplets based on their distances. This method was shown to significantly outperforms other approaches. The idea is to draw samples uniformly according to their relative distance from one another. Such sampling can correct the bias induced by the geometry of the embedding space while still visiting every data point in the dataset. Furthermore, the method introduced a modified loss by making the margin term a function of the anchor class and learning it with the embedding function.

\subsection{Lifted structured embedding}
A more complex sample selection strategy was introduced in \cite{oh2016deep} where within a batch each anchor-positive distance is compared against all anchor-negative distances weighted by the margin constraint violation. The goal is to replace the hinge-based loss with a differentiable smooth loss using exponential weighting.
\begin{equation} \label{eq:lifted_struct}
\begin{split}
	L_{\text{lifted}} = \frac{1}{2 \left | \mathcal{P}  \right |}\sum_{(i,j) \in \mathcal{P}}\left [ \log  \left ( \sum_{(i,k) \in \mathcal{N}} \exp \{M - d(\mathbf{x}_{i},\mathbf{x}_{k}) \} + \right. \right.\\
	\left. \left. \sum_{(j,l) \in \mathcal{N}}  \exp \{M - d(\mathbf{x}_{j},\mathbf{x}_{l}) \} \right ) + d(\mathbf{x}_{i},\mathbf{x}_{j})    \right ]_{+}^{2}
\end{split}
\end{equation}
where $\mathcal{P}$ is a the set of positive and $\mathcal{N}$ is the set of negative pairs. This loss function has minimal computation overhead compared to online semi-hard mining and requires no extra change to the batch sampling algorithm thus making it easy to adopt.  However, the method still requires  large batch sizes as semihard mining.

\subsection{N-pairs embedding}
The idea of a smoother loss function was taken one step further by Sohn et al. \cite{sohn2016improved} along with a more efficient batch composition strategy which samples pairs of images from $N$ unique classes. The proposed loss function computes softmax cross-entropy on the pairwise distances within each batch.
\begin{dmath} \label{eq:npairs}
	L_{\text{npairs}} = \frac{1}{ \left | \mathcal{B}  \right |} \sum_{a \in \mathcal{B}} \left \{ \log  \left [ 1 + \sum_{ \substack{n \in \mathcal{B} y_{n} \ne y_{a}, y_{p}} } \exp (d(\mathbf{x}_{a},\mathbf{x}_{n}) - d(\mathbf{x}_{a},\mathbf{x}_{p}))  \right ] \right \}
\end{dmath}
where $ \mathcal{B}$ is the batch, and $y_{i}$ is the label of sample $i$. This batch composition strategy allows for further variations to be introduced with the hope of extracting more useful positive and negative pairs from the same batch. One such variation takes the average of n-pairs loss with the pairs reversed. Another strategy samples a few pairs from the same image with different augmentation.

\subsection{Angular loss}
Since the N-pairs method effectively solves the smoothness problem along with the need of large batch sizes it became a popular testbed for future improvements. One such improvement was introduced by Wang et al \cite{wang2017deep} which operates by constraining the angle at the negative point of the anchor-positive-negative triangle. 
\begin{equation}
{\scriptstyle
	L_{\text{angular}} = \frac{1}{ \left | \mathcal{B}  \right |}\sum_{a \in \mathcal{B} } \left \{ \log \left [ 1 + \sum_{\substack{n \in \mathcal{B} \\ y_{n} \ne y_{a}, y_{p} } } \exp (f_{a,p,n}) \right ] \right \}
}
\end{equation}
where
\begin{equation}
f_{a,p,n} = 4 \tan^{2} \alpha (\mathbf{x}_{a} + \mathbf{x}_{p})^{T}\mathbf{x}_{n} - 2 ( 1 + \tan^{2} \alpha) \mathbf{x}_{a}^{T}\mathbf{x}_{p}
\end{equation}
This change introduced scale invariance on the embeddings and provided better convergence. To achieve state of the art performance this method has to be combined with N-pairs loss.

\subsection{Ranked List Loss}

DML methods mentioned above learn the embedding space by pulling all data points of the same class as close as possible. As a result they do not consider inter-class variations which results in a loss of structure in the learned space. 
Wang et al.\cite{Wang_2019_CVPR} proposes a new loss function called ranked list loss (RLL) that learns a hyper-sphere for each class instead of a single point by forcing distance between positive pair to be smaller than a constant margin (diameter of hypersphere). In RLL, for a query $x_a$, all other data points are ranked according to their similarities to the query. In each ranked list considers $N_{c} - 1$ positive points within a class $c$ as $P_{c,i}$ and $\sum_{k \neq c} N_{k}$ in negative set as $N_{c,i}$. 

$L_P$ is minimized to pull non-trivial positive points together and learn a class hyper-sphere: 
\begin{equation}
L_{\mathrm{P}}\left(\mathbf{x}_{i}^{a} ; f\right)=\frac{1}{\left|\mathbf{P}_{a, i}\right|} \sum_{\mathbf{x}_{j}^{a} \in \mathbf{P}_{a, i}} L_{\mathrm{m}}\left(\mathbf{x}_{i}^{a}, \mathbf{x}_{j}^{a} ; f\right)
\end{equation}

Similarly, the non-trivial negative points are pushed beyond the boundary $\alpha$ by minimizing: 
\begin{equation}
L_{\mathrm{N}}\left(\mathbf{x}_{i}^{a} ; f\right)=\sum_{\mathbf{x}_{j}^{k} \in\left|\mathbf{N}_{c, i}^{*}\right|} \frac{w_{i j}}{\sum_{\mathbf{x}_{j}^{k} \in\left|\mathbf{N}_{c, i}^{*}\right|} | w_{i j}} L_{\mathrm{m}}\left(\mathbf{x}_{i}^{c}, \mathbf{x}_{j}^{k} ; f\right)
\end{equation}

Both positive and negative objectives are jointly optimized with $\lambda$ controlling the balance between the sets. 
\begin{equation}
L_{\mathrm{RLL}}\left(\mathbf{x}_{a} ; f\right)=L_{\mathrm{P}}\left(\mathbf{x}_{a} ; f\right)+\lambda L_{\mathrm{N}}\left(\mathbf{x}_{a} ; f\right)
\end{equation}
where $L_{m}$ is a pair-wise constraint, $\alpha$ and $\alpha - m$ are boundaries $m$ as margin between boundaries. 
\begin{equation}
L_{\mathrm{m}}\left(\mathbf{x}_{i}, \mathbf{x}_{j} ; f\right)=\left(1-y_{i j}\right)\left[\alpha-d_{i j}\right]_{+}+y_{i j}\left[d_{i j}-(\alpha-m)\right]_{+}
\end{equation}

\subsection{Structured clustering}
Another direction in DDML is to optimize the clustering quality directly. Sohn et al. \cite{oh2017deep} proposed a framework that takes the global embedding structure into account using the facility location function \cite{Lin:2012} and optimizes the normalized mutual information (NMI). The measure of quality of a given clustering can be formulated as follows:
\begin{equation}
{\scriptstyle
	\widetilde{F}(X) = \sum_{k}^{\left | \nu \right |} \max_{j \in \{y_{i}=k\}} F(X_{ \{y_{i}=k\}}, \{j\})
}
\end{equation}
where $X$ is a set of inputs and $F$ is the facility location function. The goal is to have the clustering score greater than the maximally violating cluster assignment via the following loss:
\begin{equation}
{\scriptstyle
	L_{\text{struct. clust.}} = \left [ \max_{\substack{S \subset \nu \\ \left | S \right | = \left | \nu \right |}} \left \{ F(X,S) + \gamma \Delta (g(S), Y) \right \}  - \widetilde{F}(X) \right ]_{+}
}
\end{equation}
where $Y$ is the labels of $X$, $\nu$ is the ground truth set, $\Delta (A, N)$ is the structured margin defined as $1 - NMI(A, B)$, and $g(S)$ is a mapping function of indices to nearest clusters labels:
\begin{equation}
g(S)[i] = \argmin_{j} \left \| X_{i} - X_{\{j | j \in S\}}  \right \|
\end{equation}
The challenge with this approach is that in order to get the gradients of the loss, one must compute the solution to the first part of the loss function. Even thought the authors propose a better approach than the greedy algorithm, the method is still computationally expensive during training.

\subsection{Prototypical loss}
Prototypical networks \cite{snell2017prototypical} try to learn an embedding by diverging from the notion of positive and negative samples. Instead, the loss is formulated as a soft-nearest neighbor relationship to the correct class prototypes. Here, the prototypes are equivalent to proxies or anchors in the other formulations. Training episodes are formed randomly by first sampling a subset of classes and then for each class, sampling images to estimate a prototype and  estimate the membership of  samples to the correct class-prototype. This places a limitation on the structure of a batch and can make scaling across GPU's harder.
\begin{equation}
{\displaystyle
	L_{proto} = - \log \bigg(\frac{\exp(-d(\mathbf{x}_{a},\mu_a)))}{ \sum_{k\in \mathcal{K}}\exp(-d(\mathbf{x}_{a},\mu_k)) }\bigg)
},
\end{equation}
where $\mathcal{K}$ represents the classes present in an episode.
\subsection{Proxy-based loss}
Similar to prototypical networks, the idea of proxy-based losses is to replace positive and negative samples with points that represent the ideal cluster center of each class. In this case however, these class centers are called proxies as they are initialized randomly and learned along with the embedding function. In theory, every triplet-based loss can be transformed into using proxies. Movshovitz-Attias et al. proposes the use of exponential weighting of the distances using the proxy-based NCA loss \cite{movshovitz2017no}:
\begin{equation}
{\displaystyle
	L_{pNCA}(x_a) = - \log \bigg(\frac{\exp(-d(\mathbf{x}_{a},p(a)))}{ \sum_{n\in \mathcal{N}}\exp(-d(\mathbf{x}_{a},p(n))) }\bigg)
}
\end{equation}
where $p(x)$ is the proxy of sample $x$ which is typically statically assigned before training. We will test proxies with triplet loss and NCA loss as well to highlight the differences in performance. A further improvement was introduced by Zhai et. al \cite{normproxies} by adding layer and weight normalization to the penultimate layer and computing the softmax over the cosine distances instead of NCA.
\begin{equation}
L_{pSoftmax} = -\log \bigg(\frac{\exp (\mathbf{x}_{a}^{T}p(a)  )}{\sum_{k \in P} \exp(\mathbf{x}_{a}^{T}p(k))} \bigg)
\end{equation}
where $P$ is a set of all proxies. This approach needs a very large embedding size and strong regularization (e.g. dropout) to avoid over-fitting. The authors noticed that these large embeddings are sparse, and due to layer and weight normalization can be thresholded at 0 into binary to use less total number of bits without much loss in performance. In fact, a 2048 dimensional binary embedding requires the same number of bits as a 64 dimensional float embedding so nearest neighbor computations are comparable.

\subsection{Ensemble methods}

Ensemble of weaker models is a popular approach to get a performance boost \cite{breiman2001random,friedman2001greedy}. Naturally, ensemble learning has also been explored in DML domain. Few recent examples are BIER\cite{opitz2017bier}, HDC\cite{yuan2017hard} and DREML\cite{Xuan_2018_ECCV}. We consider DREML in our evaluation as the most recent ensemble approach with the highest reported performance. In DREML, the authors create a collection of related models each of which learn an embedding. Each model sees a subset of data partitioned on class labels, the final embedding is derived by concatenating each independent model embedding. The hope is that by combining several high-bias, low variance models the resulting prediction will be low-bias and low-variance\cite{breiman1996bagging}.
While it is common knowledge that by adopting ensemble learning there is likely a performance boost, in this case it becomes difficult to discern the effects of ensemble vis-a-vie other design choices made. This became apparent to us when our implementation of DREML without ensemble learning is not close to the reported performance (Table \ref{tab:CUB-inception}, \ref{tab:CARS-inception}). 

\subsection{Other methods}
There are several other methods for DDML that are worth mentioning, but are excluded from this study. Recent research focuses on improving the retrieval performance using boosting \cite{BIER} or attending diverse spatial locations \cite{WonsikAttentionEnsemble}. These methods achieve higher recall by using a much large embedding space. Other approaches maintain a hierarchical relation among samples during training \cite{hierarchical_triplet}. Though this approach achieves good performance, computing a pair-wise distance matrix on the whole dataset is infeasible in practice. Lastly, clustering quality can also be optimized directly via relaxing the problem of clustering with Bergman divergences \cite{spectral_clustering} to improve the structured clustering loss. However, this method requires very large batch sizes and is computationally expensive.


\section{Datasets}
For evaluation we choose the following public image datasets. In all cases the test set is both the query and index set.

\textbf{CUB-200-2011} \cite{WahCUB_200_2011}  features 11,788 images over 200 classes of birds. We followed the standard splits by using the first 100 classes for training and the remaining classes for testing.

\textbf{CARS-196} \cite{cars196} contains 16,185 images over 196 classes of various cars. The first 98 classes (8,054 images) were used for training, the remaining 98 classes (8,131 images) for testing.

\textbf{Stanford Online Products} \cite{oh2016deep}  features 120,053 images over 22,634 classes. 11,318 classes with 59,551 images are used for training and the other 11,316 classes with 60,502 images are used for testing. This dataset is excellent for testing the scalability of various methods over many classes with few images each. Due to batch-size constraints we omitted the experiments with triplet semi-hard and lifted structured loss on this dataset.

\section{Experiments}
We followed the same evaluation protocol as in \cite{oh2016deep} by computing the clustering quality using NMI and retrieval performance by Recall@K. We measured Recall@K by first computing every embedding in the test set. For each embedding we retrieved the nearest K neighbors in the embedding space using the Euclidean distance. If at least one embedding in the retrieved set had the same label as the query we assigned a score of 1, otherwise 0. The final Recall@K is the mean of these scores over the whole test set. The main goal with our experiments was to test every method under fair circumstances which involved grid search to find the best performing hyper-parameters. These parameters are published along the codebase.

\begin{table*}[t!]
	\centering
	\begin{tabular}{@{}l|llllll@{}}
		\toprule
		Recall @K & \multicolumn{1}{c}{1} & \multicolumn{1}{c}{2} & \multicolumn{1}{c}{4} & \multicolumn{1}{c}{8} & \multicolumn{1}{c}{16} & \multicolumn{1}{c}{NMI} \\ \midrule
		Triplet Semihard & 50.9 | 48.2 & 63.3 | 60.9 & 74.8 | 72.1 & 84.3 | 82.1 & 91.1 | 89.6 & 61.1 | 60.0 \\
		Lifted Struct  & 51.7 | 50.0 & 63.2 | 62.8 & 74.6 | 74.0 & 83.8 | 83.3 & 90.6 | 90.3 & 60.9 | 60.5 \\
		N-Pairs & 53.7 | 56.1 & 65.9 | 68.2 & 76.4 | 78.8 & 85.2 | 87.4 & 91.9 | 92.9 & \underline{63.6} | 65.5 \\
		Struct Clust & \underline{56.4} | - & \underline{67.8} | - & \underline{78.2} | - & \underline{86.5} | - & \underline{92.2} | - & \textbf{64.0} | - \\
		Margin Loss & 51.3 | \underline{59.6} & 63.2 | \underline{71.0} & 74.3 | \underline{81.0} & 83.4 | \underline{88.2} & 90.4 | \underline{93.1} & 61.1 | \underline{67.3} \\
		Angular Loss & 53.2 | 59.1 & 66.2 | 71.6 & 76.4 | 81.2 & 85.5 | 88.2 & 91.5 | 93.2 & 62.8 | 66.3 \\
		Prototype Loss & 47.8 | 54.4 & 60.9 | 67.2 & 73.0 | 78.0 & 83.3 | 86.2 & 90.6 | 92.6 & 60.7 | 64.8 \\
		Proxy-Triplet & 50.5 | 53.1 & 62.4 | 65.3 & 73.4 | 76.1 & 82.6 | 84.5 & 89.5 | 90.8 & 59.6 | 62.9 \\
		Proxy-NCA & 54.6 | 58.1 & 66.5 | 70.0 & 77.0 | 79.1 & 85.7 | 86.3 & 91.7 | 91.8 & 63.2 | 64.1 \\
		Proxy-Softmax & \textbf{58.3} | \textbf{63.5} & \textbf{70.4} | \textbf{74.3} & \textbf{80.5} | \textbf{82.6} & \textbf{88.6} | \textbf{89.7} & \textbf{93.4} | \textbf{94.1} & \textbf{64.0} | \textbf{69.5} \\
		Ranked List Loss & 51.4 | 50.2 & 64.0 | 62.7 & 74.5 | 73.6 & 84.2 | 83.1 & 90.1 | 89.6 & 61.8 | 58.5 \\
		DREML$\dagger$ & 55.4 | 59.0 & 67.0 | 71.0 & 77.1 | 80.5 & 85.6 | 87.5 & 91.1 | 92.7 & 61.0 | 63.2 \\
	\end{tabular}
    \vspace{.1em}
	\caption[]
	{\tabular[t]{@{}l@{}}Recall and NMI scores on the CUB200 dataset with Inception-BN (first column) and ResNet50 (second column) \\ $\dagger$: Second column is with ResNet18\endtabular}
	\label{tab:CUB-inception}
\end{table*}

\subsection{Implementation}
We used  MXNet \cite{chen2015mxnet} v1.4 to as our framework. We tested GoogleNet \cite{szegedy2016rethinking} with batch normalization \cite{batchnorm}, and Resnet50 \cite{resnet} backbones pre-trained on ImageNet with each method. We also tested different embedding and batch sizes. The output of the final layer was normalized where indicated in Table \ref{experiment-variety}. We used the Adam optimizer for all experiments except for Structured Clusters where we used rmsprop \cite{tieleman2012lecture} with an exponentially decayed gamma factor. The batch size was kept the same for all method at 120, embedding size at 64 and used a single Nvidia Tesla V100 GPU. We used an embedding size of 2048 for proxy methods with softmax loss thresholded at 0 to match 64 dimensional float embeddings. For DREML we used NCA loss and a final embedding size of 144 (L=12, D=12) which we found as the smallest reasonable setup (with still larger embeddings compared to all other methods). During training, images were resized to 256x256 then a 224x224 crop was sampled with 50\% chance of horizontal flipping. At test time, we use only the middle crop of the original image. 

\subsection{Results and Discussion}
We summarize the results of all algorithms on the three datasets in contrast to the corresponding reported score in table \ref{tab:CUB-inception}, \ref{tab:CARS-inception}, and \ref{tab:SOP-inception}. Surprisingly, we find that most methods perform a lot better than expected. In particular, losses reported with GoogLeNet yield much higher recall and NMI with the added batch normalization. On the other hand, margin loss performs similar to NPair loss unless the number of classes is high where distance weighted sampling can shine. The best performing algorithm in terms of retrieval is dominantly the normalized proxies trained with cross-entropy loss. Interestingly, though structured clustering consistently produce the highest NMI it falls behind in terms of retrieval suggesting strong class entanglements.

In the following we summarize the learnings for method along with some practical recommendations.

\textbf{Triplet semi-hard and lifted structures}: These methods do not solve the sampling problem, converge slow, and require very high batch sizes, especially when the number of classes is large. Our evaluation is in line with the related results, in fact every other algorithm performed better on all datasets. Thus, we consider future comparisons against triplet semi-hard loss obsolete and unnecessary.

\textbf{N-Pairs} is a very stable easy to implement algorithm with an average performance that can provide excellent baselines for future research without a need for large batch sizes. Moving from the common triplet (semi-hard) loss future baselines should start with N-Pairs loss. Interestingly larger embedding sizes did not yield better results, even though it uses softmax-based features similar to the proxy-softmax method.

\textbf{Structured Clusters} is a very unstable algorithm that is very sensitive to hyper-parameters. We found that training stability strongly depends on the batch composition: too many unique classes can destabilize training while too few slows the process down. Coupled with the much larger runtime complexity and complex implementation this algorithm is an \textit{interesting bird} but unfit for practical applications even though it has the best clustering performance in our analysis.

\textbf{Margin Loss} performs relatively well with ResNet50 backbone, but not nearly as good as expected with Inception-BN. Most likely this can be attributed to the per-class margin which benefits from better feature extractor. The sampling strategy is also slow and does not scale well with the number of classes.

\textbf{Angular Loss} is one of the best non-parametric loss function in terms of retrieval. Even though the original paper recommends to use it together with N-Pair loss we found no improvement in that setup. In fact, we found that angular loss performs worse on the Standford Cars and Online Products datasets if the embedding size is 64. This is in contrast to the original paper which states that the embedding size has no effect on the performance.

\textbf{Prototypical Loss} is easy to train and the sampling schemes are easy to implement. There is more room for improvements in this algorithm as it was mostly conceived as a few-shot method as opposed to a DML approach\cite{ravinash2019}. Training long and slow is a key for obtaining good performance from this algorithm. The strategy which gives the best performance boost is to use more than one episode per batch or aggregate the gradients over multiple batches before back propagation.

\textbf{Proxies}: NCA-based loss outperforms the triplet variant, but it needs a few tricks to avoid early over-fitting. Since proxies and embeddings are normalized, training can stall when relative distances become very small. One way to solve this problem is to add a sufficiently small temperature parameter or to scale the embeddings with a constant factor. We found that without the optimal scaling proxy-based methods perform very poor compared to non-parametric DML losses.

\textbf{Ranked List Loss}: Even though we managed to reproduce the published results, the performance is far worse when the embedding size was shrunk from 512 to 64 and extra model-tuning steps like multi-scale embedding layers were removed. This hints that those aids might be the main contributors for the reported high results in the original paper. On the SOP dataset, we never managed to make this algorithm converge. A possible explanation might be the low number of images per class in the dataset preventing the model to learn good hyper-spheres around class centers.


\begin{table*}
	\centering
	\begin{tabular}{@{}l|cllllll@{}}
		\toprule
		Recall @K &  \multicolumn{1}{c}{\begin{tabular}[c]{@{}c@{}}Reported\\ (@1)\end{tabular}} & \multicolumn{1}{c}{1} & \multicolumn{1}{c}{2} & \multicolumn{1}{c}{4} & \multicolumn{1}{c}{8} & \multicolumn{1}{c}{16} & \multicolumn{1}{c}{NMI} \\ \midrule
		Triplet Semihard & $\text{51.54}$\rlap{$^*$} & 60.94 & 72.93 & 82.95 & 89.57 & 94.08 & 57.55 \\
		Lifted Struct    & $\text{49.20}$\rlap{$^*$} & 65.29 & 75.75 & 83.95 & 89.87 & 94.07 & 61.00    \\
		N-Pairs           & $\text{71.12}$\rlap{$^*$} & 71.18 & 80.68 & 87.47 & 92.36 & 95.55 & \underline{63.68} \\
		Struct Clust & 58.10 & \underline{73.26} & \underline{82.41} & \underline{88.8}  & \underline{93.37} & \underline{96.21} & \textbf{64.84} \\
		Margin Loss      & $\text{\underline{79.60}}$\rlap{$^\dagger$} & 70.38 & 79.84 & 87.02 & 92.18 & 95.51 & 61.32 \\
		Angular Loss     & 71.30 & 68.86 & 78.92  & 86.35 & 91.18 & 94.65 & 59.75 \\
		Prototype Loss   & - & 59.49 & 71.86 & 82.03 & 89.45 & 94.31 & 58.44 \\
		Proxy-Triplet    & 55.90 & 65.34 & 76.01 & 84.53 & 90.96 & 94.92 & 60.85 \\
		Proxy-NCA        & 73.22 & 71.90  & 81.68 & 87.94 & 92.29 & 95.82 & 62.45 \\
		Proxy-Softmax & \textbf{81.70} & \textbf{76.98} & \textbf{85.33} & \textbf{90.83} & \textbf{94.88} & \textbf{97.36} & 61.34 \\
		Ranked List Loss & 74.00 & 69.83 & 79.66 & 87.00 & 92.19 & 95.68 & 62.28 \\
		DREML & 84.20 & 75.93 & 84.44 & 90.00 & 94.11 & 96.85 & 61.23 \\
	\end{tabular}
    \vspace{.1em}
	\caption{Recall and NMI scores on the CARS196 dataset with Inception-BN. $*$: paper used GoogLeNet, $\dagger$: paper used ResNet50}
	\label{tab:CARS-inception}
\end{table*}

\begin{table*}
	\centering
	\begin{tabular}{@{}lcllllll@{}}
		\toprule
		Recall @K &  \multicolumn{1}{c}{\begin{tabular}[c]{@{}c@{}}Reported\\ (@1)\end{tabular}} & \multicolumn{1}{c}{1} & \multicolumn{1}{c}{2} & \multicolumn{1}{c}{4} & \multicolumn{1}{c}{8} & \multicolumn{1}{c}{16} & \multicolumn{1}{c}{NMI} \\ \midrule
		N-Pairs & 67.7 & 61.24 & 67.18 & 72.36 & 77.02 & 81.27 & 67.31 \\
		Struct Clust & 67.0  & 64.27 & 69.65 & 74.33 & 78.58 & 82.41 & 86.75 \\
		Margin Loss & 72.7 & 67.63 & 73.20 & 77.98 & 82.11 & 85.54 & 87.68 \\
		Angular Loss & 70.9 & 68.97 & 74.29 & 78.75 & 82.60 & 85.99 & 87.55 \\
		Prototype Loss & - & 61.82 & 67.76 & 73.28 & 77.94 & 82.18 & 85.76 \\
		Proxy-Triplet & - & 61.41 & 67.15 & 72.31 & 77.00 & 81.31 & 86.91 \\
		Proxy-NCA & \underline{73.7} & \underline{73.56} & \underline{78.42} & \underline{82.39} & \underline{85.56} & \underline{88.21} & \underline{88.93} \\
		Proxy-Softmax & \textbf{73.8} & \textbf{74.30} & \textbf{80.13} & \textbf{82.41} & \textbf{87.62} & \textbf{89.93} & \textbf{89.10} \\
	\end{tabular}
    \vspace{.1em}
	\caption{Recall and NMI scores on the Stanford Online Products dataset with Inception-BN. Triplet-Semihard and Lifted Struct. methods were omitted due to batch size constraints. $*$: paper used GoogLeNet, $\dagger$: paper used ResNet50}
	\label{tab:SOP-inception}
\end{table*}

\subsection{Embedding Size}
Based on \cite{oh2016deep} the size of the embedding vector plays no role in the performance. This insight has been widely adopted and methods  use larger embeddings in their comparisons. Recently Zhai et al. has shown the beneficial impact of the embedding size when softmax-based features are used in the loss function \cite{normproxies}. Our results show the same trend across different algorithms (see Figure \ref{fig:embeddingsize}) hinting that hinge-based losses cannot take advantage of larger batch sizes.

\subsection{Backbone}
We investigated the effect of using a different backbone to test the effect of the feature space. Our hypothesis was that a better feature extractor will yield strictly better embedding performance across all methods. We used the ResNet50 feature extractor and compared the results against Inception-BN while keeping all the other parameters the same. Interestingly, most methods benefited very little, or actually regressed with the new backbone. The only exception was margin loss which worked much better with ResNet50, increasing R@1 from 51.3\% to 59.6\% on CUB200. A possible explanation is that ResNet50 is a better function approximator leading to overfit on such small data sets. Hence, when the available data is limited a less performing feature extractor could act as a regularizer more effectively. The final features size of ResNet50 is also double as large compared to the one from Inception-BN (2048 vs 1024) which might also play a role here.

\subsection{Sampling Strategy}
There are different sampling strategies that one could use such as the usual minibatch sampling or episodic sampling. While it is possible to replace minibatch sampling with episodic sampling the reverse is usually not possible. In our experiments, we found little evidence that this mattered. Also, for episodic sampling, we could sample with replacement or without replacement. We also found that such changes did not contribute much to the final accuracy of the methods. Hence, we use the default sampling approach recommend by each method.

\subsection{Batch size}
Due to online hard/semi-hard mining batch size plays a crucial role in the effectiveness of DML algorithms like triplet-loss or lifted structures. However, it is unclear if larger batches are similarly useful or even required for non-mining DML algorithms, especially proxy-based losses. Thus, we re-trained several methods on the previously found hyper-parameters, but with varying batch sizes between 2 and 128. Results on CUB200 are seen on Figure \ref{fig:batchsize}; we witnessed the same characteristics on the other two datasets as well.

We found that except the proxy-softmax method all other loss functions benefit from a larger batch size. This result hints that in order to learn a more compact representation (i.e.\ small embedding size) a larger slice of the global latent space is necessary to be considered during training. This also explains why the proxy-softmax method achieves a higher score even with small batch sizes. The performance of margin loss declines with larger batch sizes above 32 which we attribute to its distance-weighted sampling that acts as an effective online mining operation.

\section{Conclusions}
In this paper we explored state of the art deep distance metric learning approaches. We shared their performance numbers across standard datasets under same hyper-parameter settings which has not been done before. We have also shared our insights into why a particular approach does well in a particular set of conditions but not well in others. We found that recent deep metric learning algorithms perform very differently under conditions that were establish in earlier papers. However, our results indicate that the underlying relationship between the performance of loss functions and the models' embedding size, feature extractor, and batch size is not trivial and require further research to deepen our understanding.


{\small
\bibliographystyle{ieee}
\bibliography{amlc_dml_2019}
}

\end{document}